%% file: main.tex
\documentclass[10pt,conference]{IEEEtran}

\usepackage[english]{babel}
\usepackage[utf8]{inputenc} 
\usepackage[T1]{fontenc} 
\usepackage{xcolor}
 
\usepackage{tcolorbox}

\usepackage{microtype}
\microtypesetup{protrusion=false}

\usepackage{balance}

\usepackage{amsmath,amsfonts} 
\usepackage{graphicx}
\usepackage{algorithm}
\usepackage{algorithmic}

\usepackage{subcaption}
\usepackage{listings}

\usepackage{hyperref}
\usepackage{cleveref}
\usepackage{booktabs}
\usepackage{multirow}
\usepackage{float}

\usepackage{siunitx} 

\usepackage{mathtools}

\title{On Enhancing Root Cause Analysis with SQL Summaries for Failures in Database Workload Replays at SAP HANA}
\author{\IEEEauthorblockN{Neetha Jambigi}
\IEEEauthorblockA{University of Cologne \\
Cologne, Germany }
\IEEEauthorblockA{njambigi@smail.uni-koeln.de}
\and
\IEEEauthorblockN{Joshua Hammesfahr,\\ Moritz Mueller}
\IEEEauthorblockA{ 
SAP\\ Walldorf, Germany\\ \{joshua.hammesfahr, \\moritz.mueller07\}@sap.com
}
\and

\IEEEauthorblockN{Thomas Bach}
\IEEEauthorblockA{SAP \\ Walldorf, Germany\\
0000-0002-9993-2814}

\and
\IEEEauthorblockN{Michael Felderer}
\IEEEauthorblockA{German Aerospace Center \\ University of Cologne \\
Cologne, Germany \\
0000-0003-3818-4442}
}

\begin{document}

\maketitle

\begin{abstract}

Capturing the workload of a database and replaying this workload for a new version of the database can be an effective approach for regression testing. However,  false positive errors caused by many factors such as data privacy limitations, time dependency or non-determinism in multi-threaded environment can negatively impact the effectiveness. Therefore, we employ a machine learning based framework to automate the root cause analysis of failures found during replays. 
However, handling unseen novel issues not found in the training data is one general challenge of machine learning approaches with respect to generalizability of the learned model.
We describe how we continue to address this challenge for more robust long-term solutions. From our experience, retraining with new failures is inadequate due to features overlapping across distinct root causes. Hence, we leverage a large language model (LLM) to analyze failed SQL statements and extract concise failure summaries as an additional feature to enhance the classification process.
Our experiments show the F1-Macro score improved by 4.77\% for our data. We consider our approach beneficial for providing end users with additional information to gain more insights into the found issues and to improve the assessment of the replay results.


\end{abstract}

\section{Introduction}

\input{introduction}


\section{Failure Analysis with MIRA}
\input{problem_statement}

\input{data_evolution}
\section{Feature Extraction}
\input{methodology}

\section{Evaluation and Discussion}
\input{evaluation}

\section{Limitations and Improvements}

\input{limitations_future_work}

\section{Conclusion}
\input{conclusion}

\balance
\bibliographystyle{IEEEtran}
\bibliography{references}  
\end{document}

%% file: introduction.tex

Replaying recorded database queries poses several practical challenges, including the complexity of large replays, recording costs, legal considerations, and the occurrence of false positives. Non-deterministic factors, such as multi-threading and concurrency, randomness, and missing functionality in the recording tool, contribute to errors that do not signify regressions~\cite{jambigi2022automatic}. External factors like network and infrastructure issues further contribute to a variety of reasons for false positives in practice~\cite{bach2022testing}. This can result in up to 1 million failures per replay~\cite{jambigi2022automatic}. Deciding whether a failure is a true positive or false positive case requires to identify the root cause of an error to finally assess whether a software regression exists. However, manual root cause identification and assessment is impractical due to the large amount of failures. Previous work proposed a machine learning-based automated root cause analysis system called MIRA to attribute root causes to failures arising from a capture and replay-based testing approach~\cite{jambigi2022automatic}.
MIRA has been in productive use for capture and replay testing the database management system SAP HANA~\cite{FarberHANA2012journal, FarberHANA_SIGMOD2012}. In addition to testing for regressions, capture and replay is also used to perform evaluations of new features, identify performance issues, and other purposes. The approach is rather successful as MIRA helped to detect one-third of potential software regressions for upcoming major versions~\cite{jambigi2022automatic}.

MIRA is a supervised learning setup that relies on the features of a SQL statement collected during the replays as shown in \Cref{tab:DataTypeOverview}. MIRA attempts to learn representations for the SQL statement strings in combination with other textual attributes like error messages using \textit{TFIDF}~\cite{salton1986introduction}, \textit{Doc2Vec}~\cite{le2014distributed}, or \textit{fasttext} ~\cite{bojanowski2017enriching} in order to use them for classifying replay failures. However, as the observed software evolves, new failures arising from new software features demand continuous adaptation of MIRA to be able to identify root causes for new replay failures. With the current set of attributes, failures from several different root cause categories tend to be almost indistinguishable by classification models, due to extremely overlapping features across failure categories. Manual exploration of regressions arising during replays encompasses a comprehensive understanding of the events preceding a statement's failure during the replay process. More than 10\% of the root cause categories require this historical context to be classified under the right category. To tackle these issues arising from data evolution, we propose encoding the historical information of a failed SQL statement to further contextualize a replay failure. Previous work focus on developing effective representations for SQL queries using natural language processing techniques in order to model various aspects of databases~\cite{tang2021forecasting, jain2018query2vec, li2019detection, zahir2016recommendation}. Our approach involves constructing a new feature by utilizing a large language model (LLM) analyzing a sequence of failed SQL statements to extract failure summaries. 
 
In the following sections we provide an overview of MIRA and the application in a production environment, outline the encountered challenges, and present a method for enhancing the current setup to accommodate the evolving data landscape.


%% file: problem_statement.tex
In this section we discuss the performance of MIRA in a production environment, user feedback on its performance, and possible measures to address drifting data.
\vspace{-2mm}
\subsection{Performance in Production}
Captured workload consists typically of SQL statements. SQL statements executed during a replay are referred to as 'events'. Failed events from a replay are forwarded to MIRA and this workflow is presented as a part of \Cref{Fig:NewTrainingData}. Each failed event is assigned a root cause category along with measures to indicate the certainty of the classification. The predictions that are deemed uncertain are manually assessed and reclassified by 'operators' who are the domain experts. During the time MIRA has been in use, there have been several instances of reclassifications by operators. These include instances where additional context is required to explain the reclassification, as well as a few instances of human error, and instances that the classifier could benefit from. The classifier undergoes weekly retraining, incorporating new failure data and operator-reclassified events. Before each training cycle, cross-validation, coupled with hyperparameter optimization, is conducted, monitoring F1-Scores. A significant decline in F1-Score from the previous iteration may signify potential issues with training items with overlapping features of different classes or human errors in the reclassification process. An expert reviews these items and makes decisions concerning training data and subsequent classifier training.  The current setup in MIRA in production is a KNN~\cite{mucherino2009k} classifier with custom distance~(CD)~\cite{jambigi2022automatic} using features as shown in \Cref{tab:DataTypeOverview}.We employ KNN because of its interpretability for analyzing predictions, which is important in production settings. We concatenate and vectorize textual attributes as a single feature using \textit{fasttext}~\cite{bojanowski2017enriching} and one-hot-encode the categorical attributes. 

\begin{table}[ht]
\centering
\caption{Overview of Data Types}
\resizebox{\columnwidth}{!}{%
\label{tab:DataTypeOverview}
\begin{tabular}{llll}
\hline
Attribute             & Type(\begin{tabular}[c]{@{}l@{}}\#Values\end{tabular}) & Example\\ \hline
Error Code            & Categorical (142)              & -303, 111                  \\
Request Name          & Categorical (10)               & 1, 2                       \\
SQL Type              & Categorical (7)               & 1, 19, 5                   \\
SQL Sub Type          & Categorical (79)              & 1, 2, 3                    \\
Error Message         & Text (-)               & 'Cannot find table/view X' \\
SQL Statement Strings & Text (-)               & 'Select * from table X'    \\
\hline

\end{tabular}
}
\end{table}

Over the course of 2.5 years, over 1,000 replays from 180 unique workload captures have been processed through MIRA. Currently, the total number of failure categories has grown to 233, which is 2.5 times the original count of 93~\cite{jambigi2022automatic}. Over the span of the past 12 months, on average 30 replays per month have been processed through MIRA with an average of failures from 9 unique root causes per replay. The average number of failed statements per replay is 31,600 and the average size of a replay is over 670 million events. The latest model configuration achieved an average F1-Macro score of 82.45\% across 20 iterations during training cross-validation. Subsequently, with the introduction of flagging predictions from \textit{problem group}, which is further explained in \Cref{ss:mitigatingIssues},  the average F1-Macro score over 17  trainings has decreased to 81.48\%. 

 

\vspace{-1mm}
\subsection{Mitigating Data Evolution Issues}
\label{ss:mitigatingIssues}
Ongoing replay analysis reveals new failure types, demanding continuous adaptation of our current setup and models in MIRA. 
With the evolving data, there has been a growing set of replay failures that look identical despite considerably different root causes. The uncertainty measures cannot identify situations without additional information. There are very limited samples for new root causes in MIRA, and the oversampling of minority classes can potentially impact classes with shared vocabularies, making it further challenging. However, the introduction of a new feature into a productive system requires thorough evaluation due to potentially significant ramifications. The solution should improve classification for problem categories without hampering the overall performance.

Some challenges are relatively trivial in terms of efforts needed to identify alternative mechanisms to overcome the limitations of how we vectorize textual attributes. For instance, to address changing patterns in Error Messages, replacing the original mechanism with techniques like \textit{fasttext} works effectively. However, the identical failures with very different root causes require a more robust approach like the selection of new features or feature engineering.

 
 Stack traces could help distinguish between different types of errors, but they are not universally available for all error types in this dataset. Such failures occur infrequently, making it difficult to gather more data and conduct productive experiments. Additionally, automatically obtaining precise stack traces for a failed event is currently a complex task.
 
In contrast to stack traces, each event is linked to a SQL statement string. Our assessments suggested that statement strings can help resolve overlapping features in MIRA. We recognized the potential inclusion of SQL statement strings as an attribute, in addition to Error Messages, through experiments outlined in \Cref{tab:SSEMResultsTraining}. Additionally, we explored \textit{fasttext} as an alternative for vectorizing Error Messages resulting in improvements. We used ${\chi}^2$ ~\cite{agresti2018introduction} statistical test on both Error Messages and SQL Statement strings to identify the most important terms for classification in MIRA to inform the text preprocessing for the experiments. Concatenating SQL Statement strings with Error Messages and vectorizing them using \textit{fasttext} due to the sub-word-embeddings tends to be a more effective feature across all classes.


\begin{table}[th]
\centering
\caption{Average F1-Scores of 5-fold cross-validation (CD: Custom Distance, ED: Euclidean Distance)}
\label{tab:SSEMResultsTraining}
\begin{tabular}{llcc} 
\toprule
                                                                           &               & \multicolumn{2}{c}{KNN}  \\ 
\cmidrule(l){3-4}
Feature                                                                    & Vectorization & CD    & ED               \\ 
\midrule
\multirow{2}{*}{\begin{tabular}[c]{@{}l@{}}Error \\ Message\end{tabular}}  & TFIDF         & 94.77 & 92.21            \\
                                                                           & Doc2Vec       & 94.41 & 91.37            \\
                                                                           & Fasttext      & 95.41 & 94.30            \\ 
\midrule
\multirow{2}{*}{\begin{tabular}[c]{@{}l@{}}SQL \\ Statements\end{tabular}} & TFIDF         & 93.72 & 88.68            \\
                                                                           & Doc2Vec       & 97.01 & 95.32            \\ 
\midrule
\multirow{2}{*}{Both}                                                      & TFIDF         & 94.15 & 89.90            \\
                                                                           & Doc2Vec       & 94.51 & 92.25            \\
\bottomrule
\end{tabular}
\end{table}

In spite of the inclusion of SQL Statement string for classification, as the set of failures increased, instances from classes not prone to misclassification were observed to be misclassified. This indicates a deterioration in the model's quality due to changes in the training data.  Given the set of features as shown in \Cref{tab:DataTypeOverview}, more than 20\% of failure categories involve instances that exhibit notable vocabulary overlap with instances from other classes.  This led to the identification of problematic classes. Identification of these problematic classes involves TFIDF vectorization of Error Messages and SQL Statement strings, calculating cosine similarity between vectors. A class is considered problematic if it shares training instances with over 0.95 cosine similarity with instances from other classes, forming a reference \textit{'problem group'}. If the classifier KNN, opts for one of the training items from the problem group as one of the K neighbors to classify a failed event in production, the prediction is flagged for manual inspection regardless of the certainty. There were 62,000 events that belong to this \textit{problem group} over 6 months amounting to 1.56\% of the total failed events. Correct yet uncertain predictions of MIRA are deemed false uncertainties, which are manually investigated. Adjusting certainty thresholds to minimize these may lead to overlooking potential new failures. False uncertainties, combined with predictions from problematic groups, amplify manual labor. Given the replay sizes, manual inspection may not remain a sustainable solution.


\subsection{Feedback from End-Users}

\begin{table}[ht]
\centering
\caption{Rating Definition (*: All cases)}
\label{tab:RatingDefinition}
\resizebox{\columnwidth}{!}{%
\begin{tabular}{@{}lll@{}}
\toprule
 & Certain & Uncertain \\ \midrule
Known class & \begin{tabular}[c]{@{}l@{}}Correctly classified $\rightarrow$ No Action Needed (1)\\ Misclassified $\rightarrow$ Reclassified by Operator (3)\end{tabular} & $* \rightarrow$ Reclassified by Operator (2) \\
New class & $* \rightarrow$ Manually detected by Operator (4) & \begin{tabular}[c]{@{}l@{}}New failure $\rightarrow$ \\ Assessed by Operator (1)\end{tabular} \\ \bottomrule
\end{tabular}%
}
\end{table}

\begin{figure}[ht]
 \vspace{-3mm}
    \makebox[\columnwidth][c]{
	 \includegraphics[scale=0.69]{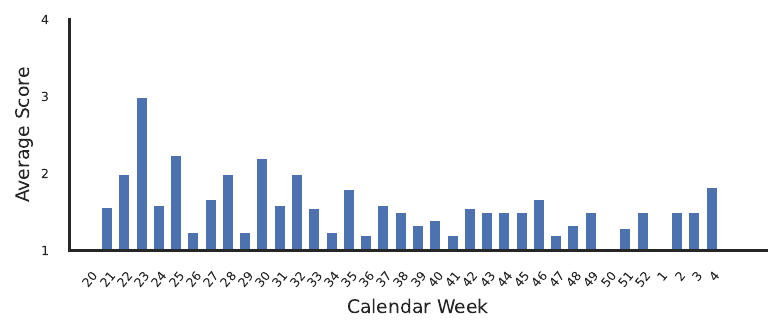}}
	 \label{datacuration}
\vspace{-4mm}
 \caption{Weekly Average User-Review}
 \label{fig:AvgUserFeedback}
 \end{figure}

The workflow of MIRA necessitates operator involvement for feedback and correction, posing an initial overhead. Following the initial learning period, the utilization of MIRA appeared to remain consistent among operators. We collected operator assessments of MIRA's results over 9 months in a production environment. As an assessment, operators provided numerical ratings per replay to indicate the severity of deviations from the expected behavior observed in MIRA's predictions. These ratings ranged from 1 to 4, with higher numbers indicating more severe deviations, and expected behavior is rated 1. More severe deviations encompass and supersede less severe deviations for a single replay. The categorization of severity ratings is summarized in \Cref{tab:RatingDefinition}. Based on collective analysis of this feedback over 199 replays 55\% of the replays are rated 1, 37\% are rated 2, 6.5\% are rated 3 and 1.5\% are 4. Average rating per week over 37 weeks is presented in \Cref{fig:AvgUserFeedback}. Overall, there are 2 cases with a rating of 4 where new failures were missed.

%% file: methodology.tex
\label{s:createNewData}
Given the challenges outlined in \Cref{ss:mitigatingIssues}, we implement a mechanism that emulates operators' processes for identifying problematic root cause categories. This entails applying filtering conditions to collect a set of SQL statements and relevant attributes, which are subsequently summarized using an LLM.

\input{data_creation}


%% file: data_creation.tex

Operators examine statements they deem relevant for identifying a failure's root cause. In a majority of analyzed cases, the root cause can be linked to statements executed within the same session ID. The session ID serves as an identifier for the session from a client (e.g., HANA Studio) to the Database Server. While a database can run multiple parallel sessions with different IDs, a single session in HANA can execute only one transaction at a time, ensuring sequential execution of all SQL statements in that session. Since concurrency still remains a challenge in the capture and replay framework, we constrain our collection to a single session ID.

The failure of subsequent statements, prompted by a preceding failed statement in our data, can be predominantly linked to unsuccessful or unexecuted DDL SQL statements. Identifying the root cause of failures is crucial, with factors like connection errors, privilege issues, or the omission of an SQL statement due to pre-processing of the workload before re-execution being important contributors to these failures. Prior to replaying the captured workload, preprocessing addresses inconsistencies arising from transactions that start before or end after the recording. Specific SQL statements are either skipped or executed based on assigned reasons. For instance, skipping a 'CREATE TABLE X' statement may result in statements accessing this table failing with the same information 'Cannot find table/view X.' Therefore, the attribute 'Skip Reasons' contributes to contextualizing the failures in a replay.

Extracting relevant SQL statements to contextualize a failed event \empty{'E'} is limited to replay events that satisfy specific conditions: i) event - failed or skipped ii) event must have been executed in the same session ID as \empty{'E'}  iii) event must have been processed prior to \empty{'E'}. The set is further filtered to retain unique SQL statements utilizing the statement hash values. 



Relevant data can also be collected using transactional dependency graphs from the capture and replay framework, which records related database objects for each SQL statement. However, it lacks object lists for DDLs like 'CREATE' statements, resulting in incomplete information. Avoiding the manual selection criteria is challenging, and it is unclear if this method provides better contextualization compared to the rule-based collection.
 

 \subsection{Using LLM for Summarizing Failed SQL}
 \label{s: llmsummary}
After applying constraints on event selection for contextualizing the failures, the sizes of these sets range from 1 to 3,314 events, with an average of 45 failed events across 25,453 instances. Vectorizing the collected statements with mechanisms like \textit{fasttext} is feasible as we employ that solution for Error Messages and SQL statements in MIRA, but the extensive contextual data necessitates intricate preprocessing steps, training, and retraining models. However, summarizing statements using a pre-trained LLM like GPT-4~\cite{OpenAI_GPT4_2023} can help reduce noise and retain only critical parts. Furthermore, LLMs are multimodal and can process diverse input types effectively~\cite{zhao2023survey, jin2024comprehensive} allowing a single input with stack traces, SQL Statements, Error Messages, etc.

\floatstyle{plain}
\newfloat{listing}{htbp}{lop}
\floatname{listing}{Listing}
\lstset{basicstyle=\ttfamily\small,breaklines=true}

\begin{listing}[htbp]
\caption{Example Prompt and Response }
\begin{tcolorbox}[colback=white, colframe=black, fonttitle=\bfseries, boxrule=0.5pt,  title=Prompt, colbacktitle=white!60!black, halign=flush left,
left=2pt, right=2pt, top=2pt, bottom=2pt,]

Given a list of JSON objects containing SAP HANA SQL statement strings, error messages, and skip reasons, generate a concise summary. Group the SQL statements based on SQL type and execution status. Summarize each group with a parseable JSON structure like:\\
\text{[\{'statement type': SQL statement type,}\\
\text{'status': failed/ skipped,}\\
\text{'error': generic summary of all 'critical' sub-parts of}\\ 
\text{unique error message in 30 words or less,}\\
\text{'objects': comma-separated list of all objects\}]} \newline \newline
Summarize 'identical' failures within the list similarly. Strictly adhere to the summary structure and include absolutely no additional information outside the JSON. Input list of JSON:\\  
\text{[\{}
\text{'Statement String': 'CALL proc\_name',}\\
\text{'Error Code': 100,}\\
\text{'Error Message': 'Transaction rolled back ....',}\\
\text{'Skip Reason': 'Skipped'}\}, \{....\}, \{....\}]
\end{tcolorbox}

\begin{tcolorbox}[colback=white, colframe=black, fonttitle=\bfseries, boxrule=0.5pt, title= Response Summary, colbacktitle=white!60!black, halign=flush left,
left=2pt, right=2pt, top=2pt, bottom=2pt,]

\text{[\{'statement type': 'CALL',}
\text{'status': 'failed',}
\text{'error': 'Operation canceled and transaction rolled back} \\
\text{\ due to exception.', 'objects': 'ABC1, ABC2'\}},
\text{\{'statement type': 'CREATE VIEW', 'status': 'failed',}
\text{'error': 'Connection error', 'objects': 'MN1, MN2'}\}]
\end{tcolorbox}
\label{SamplePrompt}
\end{listing}

Summarizing and vectorizing SQL statements to obtain representations of the workload for machine learning has been an ongoing effort \cite{deep2022comprehensive, jain2018query2vec}. However, in our case, we aim to generate a comprehensive yet succinct summary of a selected set of SQL statements and their corresponding failures to contextualize a failed event in a replay. The summaries provide an overview of relevant events leading up to a failed event in consideration. The frequency and complexity of these events play a limited role in emulating the operators' assessment process. Therefore, summarizing enables us to extract the salient information necessary to contextualize a replay failure. For summarization, we disregard the temporal ordering of the SQL statements within this group while prompting the LLM. This approach enables us to obtain a collective summary of events that exhibit similar failures. As part of the summary, we prompt for inclusion of the list of objects. However, the LLM tends to often truncate the list of objects for brevity, particularly when there is a large number of similar failures on similar statement types. Initially, we allowed unrestricted lengths for summarizing error messages. This led to summaries containing irrelevant numbers, thread IDs, and session names. Since these elements typically relate to transaction details or specific executions, they do not help us identify the root causes of failures. To address this noise, we now specify a 30-word limit on error summaries. This limit helps us focus the summaries on the most crucial parts of the error messages. We experimented with different prompting strategies and output structures by manually verifying the generated summaries across 30 samples with known failure contexts. We selected the approach that best captured the essential aspects of the failure. We then automated summarization for the remaining dataset. We preprocess the input events to remove data like names, IP addresses, hostnames, etc., to anonymize the information. \Cref{SamplePrompt} shows an example input prompt and the response SQL failure summary generated by LLM.


Our choice of models is limited by the context lengths of the LLMs.  We tested four models: llama-7b~\cite{touvron2023llama}, GPT-3.5, GPT-4, and GPT-4-Turbo. While GPT-4 provided the best summaries for instances within a 32,000 token context length, GPT-4-Turbo could handle more instances with its 128,000 token context length. The llama models struggled to generate complete summaries and adhere to the desired summary structures. Therefore, we use the GPT-4-Turbo model to summarize collected statements. Even with a 128,000 context length, 18\% of our dataset's instances exceed this limit.




\subsection{Building New Training Data}
Capture and replay-based testing produces large result sets that are only temporarily retained for operator review due to storage and cost constraints. Over the years, MIRA's training data has evolved with new root cause identifications. However, since the original result sets from most replays were lost, we cannot obtain contextual data for events in our current training.

\begin{figure}[ht]
   \makebox[\columnwidth][c]{
	 \includegraphics[scale=0.65]{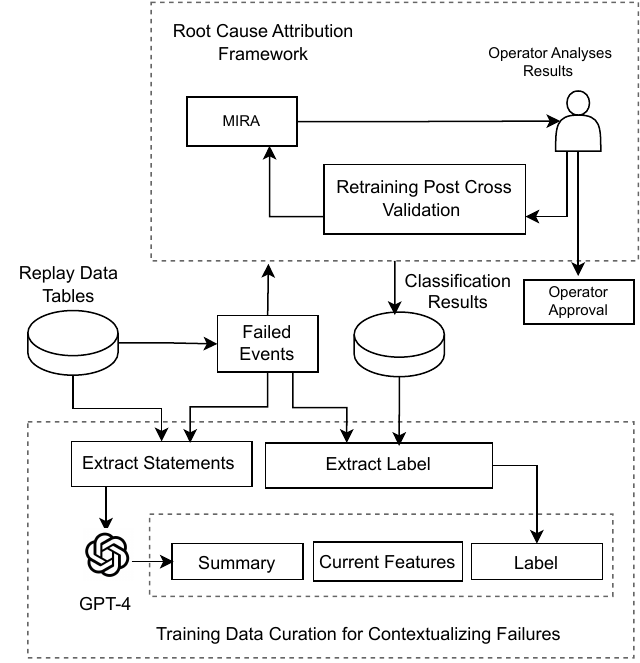}}	
 \caption{Constructing New Training Data}
 \label{Fig:NewTrainingData}
 \end{figure}

We constructed a new labeled dataset over a span of 1 year. We collect the failed events from replays and leverage the labels provided by MIRA's production setup as portrayed in \Cref{Fig:NewTrainingData}. We collect labels only for instances reclassified by an operator, indicating manual verification, or for predictions with high certainty. These conditions limit the amount of reliable data we can gather per replay. Since a single type of failure can occur multiple times in a replay, we limit samples from each failure category for a replay to 100 to ensure dataset diversity. To construct the final training dataset, we retain the same set of features as used in the productive setup, with the exception of the textual attribute. Previously, this attribute combined the Error Message and SQL Statement String. We will now enhance the textual attribute by including the summary of the contextual data for each event. 




%% file: evaluation.tex
The final dataset, after deduplication, comprises 25,453 events and 162 classes of the 233 classes from production. There is a significant class imbalance as the top 10 root causes, by count, constitute 65\% of the data. Some classes have only one sample, which prevents uniform splitting during cross-validation, even with stratification. We perform a 5-fold stratified cross-validation~\cite{stone1974cross} for hyperparameter optimization. For classifier evaluation, we present the metrics F1-Macro, F1-Comb, and Accuracy averaged across 5-fold stratified cross-validation. F1-Macro is the average F1-score of all the classes and is a suitable metric for unbalanced datasets. F1-Comb is a combined metric of F1-Macro score and certainty metrics~\cite{jambigi2022automatic} and is calculated as a harmonic mean of both. Higher F1-Comb values indicate higher certainty for correct predictions, indicating the reliability of predictions. As stated in \Cref{s:createNewData} we built a new dataset that is different than the dataset the model in production is trained on. 
We adopt a similar setting as our production environment with \textit{fasttext} to vectorize the textual attributes and \textit{KNN} as the classifier due to its interpretability. \Cref{tab:CtxtSummaryEval} presents an overview of the evaluation of concatenating the Summaries to the Error Messages and the SQL Statement strings. The F1-Macro improves by 4.77\% with the use of SQL summary and accuracy by 0.16\%.

\begin{figure}[ht]
\begin{minipage}[c][0.49\width]{0.24\textwidth}
    \subcaption{Before}
   \makebox[\columnwidth][c]{
     \includegraphics[scale=0.29]{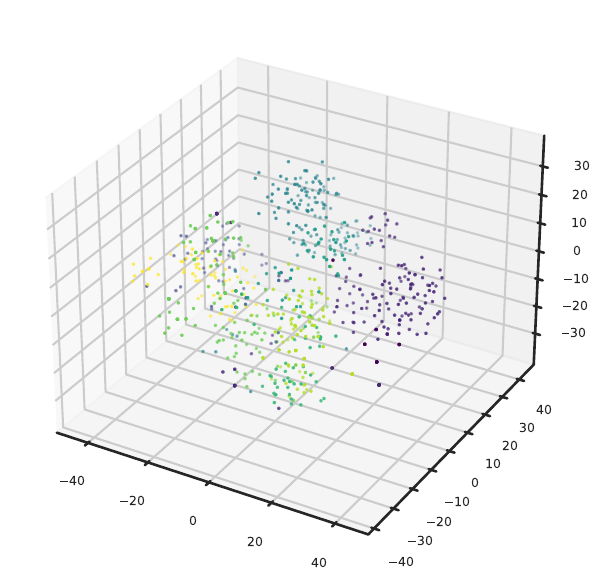}}	
     \label{NoTrace}
\end{minipage}%
\begin{minipage}[c][0.49\width]{0.24\textwidth}
     \subcaption{After}
   \makebox[\columnwidth][c]{
     \includegraphics[scale=0.29]{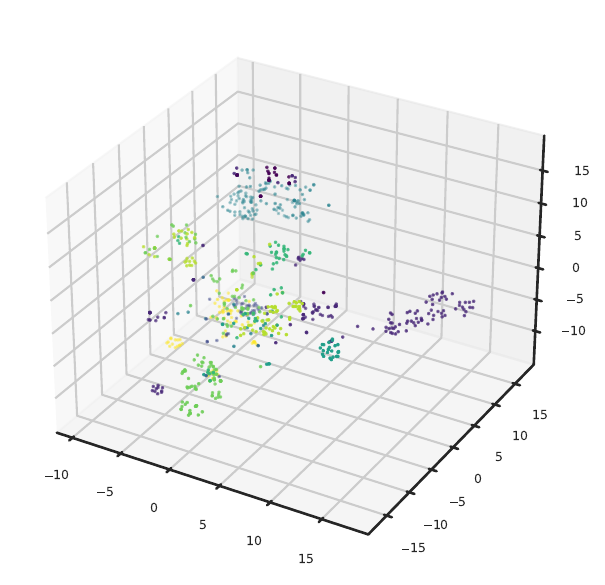}}
    \label{WithTrace}
\end{minipage}
\vspace{5mm}
\caption{t-SNE Plot - Utilizing SQL Summary}
\label{fig:tSNETextEmbeddings}
\end{figure}

 We preprocess the summaries independently prior to the concatenation of textual attributes, to retain the most informative terms using TFIDF-based term filtering. This makes the summaries further concise. Upon closer analysis, we could not find a discernible pattern to textual features in the misclassifications. However, the misclassifications largely arose from the minority classes and it is challenging to conclusively understand the issues with only a few samples. In this data, Error Messages were consistently the most important attribute for classifying root cause. To understand the impact of the summaries, we concatenated them to Error Messages and the results are comparable to the current setting of using SQL Statements and Error Messages. This makes summarized failed SQL statements a valuable feature for our classification process. Using a t-SNE plot~\cite{van2008visualizing}, \Cref{fig:tSNETextEmbeddings} visualizes the change in the \textit{fasttext} embeddings for the 10 classes with the highest frequency of failures from the \textit{problem group}. Combining Error Messages and SQL Statements with Summaries results in more cohesive clusters than using them without Summaries.






\begin{table}[tbh]
\centering
\caption{Average 5-fold cross-validation scores on new training data - Summary: Summarized failed SQL statements, EM: Error Message, SS: Statement String}
\label{tab:CtxtSummaryEval}
\resizebox{\columnwidth}{!}{%
\begin{tabular}{@{}lrrr@{}}
\toprule
Textual Attribute & F1-Comb & F1-Macro & Accuracy \\ \midrule
EM + SS & 78.82 & 63.02 & 91.86 \\
EM + SS + Summary & \textbf{79.79} & \textbf{67.79} & \textbf{92.02} \\
EM + Summary & 75.88 & 63.40 & 90.33 \\ \bottomrule
\end{tabular}%
}
\end{table}


Our data encoding involves concatenating all textual features, which may limit the impact of one feature over another. However, treating summaries as a separate feature does not significantly change the results.  While the use of summaries enhances F1-Macro by 4.77\%, it is encouraging, as this appears to be a viable feature engineering method for this data.  Further data collection and experimentation with summary structure and granularity could be beneficial to find more robust features.





\vspace{4mm}


%% file: limitations_future_work.tex

 Current constraints on data collection for contextualization presume that the root cause is present in the same session as the failed event, which may not always be accurate. Even if the LLM effectively summarizes the failed SQL statements, the necessary information for identifying the root cause might not be available in the contextual history. Furthermore, utilizing stack traces alongside failed SQL when applicable can result in a summary more informative of a root cause. Thus, an ongoing evaluation of replay failures remains essential. Additionally, GPT-4-Turbo may not have all the required domain knowledge of HANA failures to effectively summarize the failures. Finetuning or instruction tuning can be a possible solution to make the model better adapt to the domain. 
 
 
 

Prompting strategies influence LLM results~\cite{li2023prompt}. As stated in \Cref{s: llmsummary}, we designed our prompts based on evaluation across only 30 samples. To improve this process, we should gain an overall understanding of the generated summaries for the entire dataset, which could help us craft better prompts for more informative summaries.

Utilizing chained summarization~\cite{wei2022chain} to iteratively extract the top-k elements from the input can help reduce noise further and address the 18\% of the instances from our dataset that exceeded the 128,000 token context length.

Predicting future error patterns is challenging. However, we can benefit from excluding known failure patterns using the Retrieval-Augmented Generation (RAG) approach to identify mismatches that may be potential new and unknown failures.
 

 The failure summary's JSON structure allows for experimentation with subsets of the information. For example, objects from a DML SQL query may be irrelevant to other subsequent failures and can be omitted during summarization. However, validating parts of the summary requires extensive evaluation. By eliminating the JSON structure, we can free up tokens for more informative summaries.
 
We construct a new dataset for evaluating our approach, using MIRA's classification results as the new ground truth for training. However, a system that continuously adapts using manually reclassified data and employs uncertainty measures to trigger such reclassifications is susceptible to performance degradation. A thorough analysis of the new training data is essential for evaluating label validity and our results. Data collection for summarization still relies on rules needing ongoing maintenance and adaptation as new issues arise.




Operators use MIRA's UI for root cause analysis but manually examine replay data for exploring new failures. Integrating condensed SQL summaries into the UI can provide immediate insights and enhance analysis.

%% file: conclusion.tex
In this study, we address challenges arising from data evolution in our root cause analysis framework using supervised learning for replay failure analysis. As we encounter new failures, the efficacy of the machine learning framework depends upon the availability of robust data for ensuring reliable classification. We outline our efforts to improve the classification and address complexities from overlapping features in failures with different root causes. Moreover, as a long-term solution, we design an approach attempting to emulate the manual root cause analysis processes of domain experts. Leveraging LLM, we summarize a set of failed SQL statements, creating a feature that enhances our data with a potentially discriminating attribute for root cause analysis. This method improved the results by 4.77\% and the experiments enabled the development of a valuable feature for classification. The failure summaries can enhance the model interpretability and facilitate a more efficient root cause analysis process for operators.








%% file: main.bbl
\begin{thebibliography}{10}
\providecommand{\url}[1]{#1}
\csname url@samestyle\endcsname
\providecommand{\newblock}{\relax}
\providecommand{\bibinfo}[2]{#2}
\providecommand{\BIBentrySTDinterwordspacing}{\spaceskip=0pt\relax}
\providecommand{\BIBentryALTinterwordstretchfactor}{4}
\providecommand{\BIBentryALTinterwordspacing}{\spaceskip=\fontdimen2\font plus
\BIBentryALTinterwordstretchfactor\fontdimen3\font minus \fontdimen4\font\relax}
\providecommand{\BIBforeignlanguage}[2]{{%
\expandafter\ifx\csname l@#1\endcsname\relax
\typeout{** WARNING: IEEEtran.bst: No hyphenation pattern has been}%
\typeout{** loaded for the language `#1'. Using the pattern for}%
\typeout{** the default language instead.}%
\else
\language=\csname l@#1\endcsname
\fi
#2}}
\providecommand{\BIBdecl}{\relax}
\BIBdecl

\bibitem{jambigi2022automatic}
N.~Jambigi, T.~Bach, F.~Schabernack, and M.~Felderer, ``Automatic error classification and root cause determination while replaying recorded workload data at sap hana,'' in \emph{2022 IEEE Conference on Software Testing, Verification and Validation (ICST)}.\hskip 1em plus 0.5em minus 0.4em\relax IEEE, 2022, pp. 323--333.

\bibitem{bach2022testing}
T.~Bach, A.~Andrzejak, C.~Seo, C.~Bierstedt, C.~Lemke, D.~Ritter, D.~W. Hwang, E.~Sheshi, F.~Schabernack, F.~Renkes \emph{et~al.}, ``Testing very large database management systems: The case of sap hana,'' \emph{Datenbank-Spektrum}, vol.~22, no.~3, pp. 195--215, 2022.

\bibitem{FarberHANA2012journal}
F.~Färber, N.~May, W.~Lehner, P.~Große, I.~Müller, H.~Rauhe, and J.~Dees, ``The {SAP} {HANA} database -- an architecture overview,'' \emph{Bulletin of the Technical Committee on Data Engineering / IEEE Computer Society}, vol.~35, no.~1, pp. 28--33, 2012.

\bibitem{FarberHANA_SIGMOD2012}
F.~Färber, S.~K. Cha, J.~Primsch, C.~Bornhövd, S.~Sigg, and W.~Lehner, ``{SAP} {HANA} database: {D}ata management for modern business applications,'' \emph{SIGMOD Record}, vol.~40, no.~4, Jan. 2012.

\bibitem{salton1986introduction}
G.~Salton and M.~J. McGill, \emph{Introduction to modern information retrieval}.\hskip 1em plus 0.5em minus 0.4em\relax USA: McGraw-Hill, Inc., 1986.

\bibitem{le2014distributed}
Q.~Le and T.~Mikolov, ``Distributed representations of sentences and documents,'' in \emph{International conference on machine learning}.\hskip 1em plus 0.5em minus 0.4em\relax PMLR, 2014, pp. 1188--1196.

\bibitem{bojanowski2017enriching}
P.~Bojanowski, E.~Grave, A.~Joulin, and T.~Mikolov, ``Enriching word vectors with subword information,'' \emph{Transactions of the Association for Computational Linguistics}, vol.~5, pp. 135--146, 2017.

\bibitem{tang2021forecasting}
C.~Tang, B.~Wang, Z.~Luo, H.~Wu, S.~Dasan, M.~Fu, Y.~Li, M.~Ghosh, R.~Kabra, N.~K. Navadiya, D.~Cheng, F.~Dai, V.~Channapattan, and P.~Mishra, ``Forecasting sql query cost at twitter,'' in \emph{2021 IEEE International Conference on Cloud Engineering (IC2E)}, 2021, pp. 154--160.

\bibitem{jain2018query2vec}
S.~Jain, B.~Howe, J.~Yan, and T.~Cruanes, ``Query2vec: An evaluation of nlp techniques for generalized workload analytics,'' \emph{arXiv preprint arXiv:1801.05613}, 2018.

\bibitem{li2019detection}
Y.~Li and B.~Zhang, ``Detection of sql injection attacks based on improved tfidf algorithm,'' in \emph{Journal of Physics: Conference Series}, vol. 1395, no.~1.\hskip 1em plus 0.5em minus 0.4em\relax IOP Publishing, 2019, p. 012013.

\bibitem{zahir2016recommendation}
J.~Zahir and A.~El~Qadi, ``A recommendation system for execution plans using machine learning,'' \emph{Mathematical and Computational Applications}, vol.~21, no.~2, p.~23, 2016.

\bibitem{mucherino2009k}
A.~Mucherino, P.~J. Papajorgji, and P.~M. Pardalos, ``K-nearest neighbor classification,'' in \emph{Data mining in agriculture}.\hskip 1em plus 0.5em minus 0.4em\relax Springer, 2009, pp. 83--106.

\bibitem{agresti2018introduction}
A.~Agresti, \emph{An introduction to categorical data analysis}.\hskip 1em plus 0.5em minus 0.4em\relax John Wiley \& Sons, 2018.

\bibitem{OpenAI_GPT4_2023}
\BIBentryALTinterwordspacing
OpenAI, ``{GPT-4} technical report,'' \emph{ArXiv}, vol. abs/2303.08774, 2023. [Online]. Available: \url{https://arxiv.org/abs/2303.08774}
\BIBentrySTDinterwordspacing

\bibitem{zhao2023survey}
W.~X. Zhao, K.~Zhou, J.~Li, T.~Tang, X.~Wang, Y.~Hou, Y.~Min, B.~Zhang, J.~Zhang, Z.~Dong \emph{et~al.}, ``A survey of large language models,'' \emph{arXiv preprint arXiv:2303.18223}, 2023.

\bibitem{jin2024comprehensive}
H.~Jin, Y.~Zhang, D.~Meng, J.~Wang, and J.~Tan, ``A comprehensive survey on process-oriented automatic text summarization with exploration of llm-based methods,'' \emph{arXiv preprint arXiv:2403.02901}, 2024.

\bibitem{deep2022comprehensive}
S.~Deep, A.~Gruenheid, P.~Koutris, S.~Viglas, and J.~Naughton, ``Comprehensive and efficient workload summarization,'' \emph{Datenbank-Spektrum}, vol.~22, no.~3, pp. 249--256, 2022.

\bibitem{touvron2023llama}
H.~Touvron, T.~Lavril, G.~Izacard, X.~Martinet, M.-A. Lachaux, T.~Lacroix, B.~Rozi{\`e}re, N.~Goyal, E.~Hambro, F.~Azhar \emph{et~al.}, ``Llama: Open and efficient foundation language models,'' \emph{arXiv preprint arXiv:2302.13971}, 2023.

\bibitem{stone1974cross}
M.~Stone, ``Cross-validatory choice and assessment of statistical predictions,'' \emph{Journal of the Royal Statistical Society: Series B (Methodological)}, vol.~36, no.~2, pp. 111--133, 1974.

\bibitem{van2008visualizing}
L.~Van~der Maaten and G.~Hinton, ``Visualizing data using t-sne.'' \emph{Journal of machine learning research}, vol.~9, no.~11, 2008.

\bibitem{li2023prompt}
L.~Li, Y.~Zhang, and L.~Chen, ``Prompt distillation for efficient llm-based recommendation,'' in \emph{Proceedings of the 32nd ACM International Conference on Information and Knowledge Management}, 2023, pp. 1348--1357.

\bibitem{wei2022chain}
J.~Wei, X.~Wang, D.~Schuurmans, M.~Bosma, F.~Xia, E.~Chi, Q.~V. Le, D.~Zhou \emph{et~al.}, ``Chain-of-thought prompting elicits reasoning in large language models,'' \emph{Advances in Neural Information Processing Systems}, vol.~35, pp. 24\,824--24\,837, 2022.

\end{thebibliography}
